\documentclass[final,5p,times,twocolumn]{elsarticle}

\usepackage{graphicx}             
\usepackage{stfloats}             
\usepackage{booktabs}             

\usepackage{amssymb}
\usepackage{amsmath}
\usepackage{hyperref}
\usepackage{CJKutf8}
\usepackage{multirow}
\usepackage{float}
\usepackage{xcolor}
\usepackage{setspace}
\usepackage{fancyhdr} 
\usepackage{indentfirst}

\makeatletter
\def\@fnsymbol#1{\@arabic{#1}}  
\makeatother

\usepackage{mathptmx} 

\begin{document}

\begin{frontmatter}

\title{Validating the Effectiveness of a Large Language Model-based Approach for Identifying Children's Development across Various Free Play Settings in Kindergarten}


\author[label1]{Yuanyuan Yang\fnref{fn1}} 
\author[label1]{Yuan Shen\fnref{fn2}}
\author[label1]{Tianchen Sun\fnref{fn3}}
\author[label1]{Yangbin Xie\fnref{fn4}}
\affiliation[label1]{organization={Research Center for Data Hub and Security, Zhejiang Lab}, 
            city={Hangzhou},
            postcode={311121}, 
            state={Zhejiang},
            country={China}}
\fntext[fn1]{44.yangoo@gmail.com}
\fntext[fn2]{yeriishen@gmail.com}
\fntext[fn3]{Corresponding author, robertsuntc@gmail.com}
\fntext[fn4]{xieyangbin@zhejianglab.org}

\thispagestyle{empty}
\begin{abstract}
Free play is a fundamental aspect of early childhood education, supporting children's cognitive, social, emotional, and motor development. However, assessing children's development during free play poses significant challenges due to the unstructured and spontaneous nature of the activity. Traditional assessment methods often rely on direct observations by teachers, parents, or researchers, which may fail to capture comprehensive insights from free play and provide timely feedback to educators. This study proposes an innovative approach combining Large Language Models (LLMs) with learning analytics to analyze children's self-narratives of their play experiences. The LLM identifies developmental abilities, while performance scores across different play settings are calculated using learning analytics techniques. We collected 2,224 play narratives from 29 children in a kindergarten, covering four distinct play areas over one semester. According to the evaluation results from eight professionals, the LLM-based approach achieved high accuracy in identifying cognitive, motor, and social abilities, with accuracy exceeding 90\% in most domains. Moreover, significant differences in developmental outcomes were observed across play settings, highlighting each area's unique contributions to specific abilities. These findings confirm that the proposed approach is effective in identifying children's development across various free play settings. This study demonstrates the potential of integrating LLMs and learning analytics to provide child-centered insights into developmental trajectories, offering educators valuable data to support personalized learning and enhance early childhood education practices.

\end{abstract}


\begin{keyword}
Large Language Models (LLMs) \sep Learning Analytics \sep Early Childhood Education (ECE) \sep Child Development \sep Free Play 

\end{keyword}
\end{frontmatter}

\thispagestyle{fancy}

\section{Introduction} \label{sec_intd}
\subsection{Free Play in Early Childhood Education (ECE)}
\label{subsec_play}
Free play refers to \textit{a voluntary, unstructured, and child-directed activity that allows children to take charge of their play, using their imagination and creativity without adult interference} \citep{caillois2001man}. It is a fundamental aspect of childhood, intrinsic to the nature of children. ECE policies from many countries incorporate play as an essential part of children's daily activities, such as the Early Learning and Development Guidelines for Children published by \cite{china2012early}, the Early Years Learning Framework proposed by \cite{aus2022early}, and the Early Years Foundation Stage framework suggested by \cite{uk2023early}. 

Despite broad recognition by policymakers and educators, implementing free play in ECE settings remains challenging. Free play lacks specific academic goals and involves high child autonomy with flexible content and rules, making integration into structured environments difficult. Teachers often struggle with limited time for planning and managing free play \citep{Nordin2024Exploring}, and not all can supervise every outdoor activity \citep{Kurnia2024Free}. Moreover, a lack of awareness among parents and administrators about the value of play-based learning adds to the difficulty \citep{Haile2024Play-Based}. At the same time, early childhood is a period of rapid development in language, social skills, and self-awareness, requiring ongoing monitoring of each child's growth. This is particularly complex in dynamic group play settings.

Consequently, effectively integrating free play into ECE settings requires overcoming significant barriers, including practical, pedagogical, and institutional challenges. This highlights the need for innovative solutions, such as technological advancements, to support educators in facilitating and monitoring free play while maximizing its developmental benefits.

\subsection{The Application of Large Language Models (LLMs) in Education}
\label{subsec_llm}
In recent years, LLMs have achieved significant breakthroughs in the field of natural language processing. 
These models, such as OpenAI’s GPT series \citep{achiam2023gpt}, Google's BERT \citep{devlin2018bert}, Alibaba's Qwen \citep{bai2023qwen}, implicitly represent a vast amount of general knowledge from the training corpus in their billions or trillions of parameters, enabling them to solve various tasks like machine translation \citep{lample2019cross}, text summarization \citep{rothe2020leveraging} and conversation \citep{zhang2019dialogpt}. 

LLMs’ exceptional capabilities in understanding and generating content have opened new avenues for innovation across various fields, including education. A growing body of research has explored the potential applications of LLMs in educational settings. Some studies used LLMs for the generation and design of educational content \citep{kasneci2023chatgpt,abdelghani2022gpt,gabajiwala2022quiz}, many other researches applied LLMs in language learning, such as teach business English writing courses \citep{kim2023study}, facilitate students' multimodal writing process \citep{liu2024investigating}, and serve as interactive chatbots for language practice \citep{kohnke2023chatgpt}. 

These studies collectively demonstrate the potential of LLMs to transform educational practices by automating content creation, personalizing learning experiences, and enhancing student engagement. In ECE setting, these capabilities could be particularly valuable in addressing key challenges such as effectively implementing free play, providing educators with tools to monitor developmental progress, offering individualized feedback, and supporting the seamless integration of play-based learning into structured environments. However, as \cite{lan2024teachers} said, teachers are exploring ways to use AI to enhance student learning, reduce their workload, and improve teaching quality, but finding the right balance and methods for integration remains challenging. Hence, there is a pressing need for more empirical research to investigate the practical applications and validation of LLMs in educational settings to ensure their effective and equitable integration.

\section{Literature Review}
\label{sec_lr}

\subsection{The Developmental Benefits of Free Play for Children}
\label{sec_lr_1}
Numerous empirical studies have demonstrated the positive effects of free play on various aspects of children's development. \cite{razak2018impact} found that increasing outdoor free play in childcare significantly boosts children's moderate-to-vigorous physical activity (MVPA). \cite{kukkonen2020creative} argue that free play enhances young children's creative collaboration and artistic expression through shared meaning-making in open-ended group drawing activities. \cite{ellis2021oh} highlighted the benefits of woodland free play, showing that interactions with natural elements and peers promote creativity, autonomy, and social skills.

However, the efficacy of free play has been debated. Some studies suggest that structured play may yield more targeted benefits. For instance, \cite{tortella_comparing_2019} reported that partly structured play increased physical activity more effectively in 5-year-olds. Similarly, \cite{tortella_effects_2022} found greater improvements in motor skills, such as balancing, among children engaged in partly structured play. \cite{palmer_effect_2019} also observed faster motor skill development in children who participated in a 5-week structured intervention compared to those in a free play group.

Free play emphasizes child autonomy and lacks predetermined goals, which may lead to less focused outcomes compared to structured play that targets specific developmental areas. However, while structured play often supports development in one or a few specific domains, free play holds the potential to foster more holistic growth across cognitive, motor, emotional, and social dimensions. Nevertheless, most empirical studies focus on isolated developmental domains, largely due to the complexity involved in measuring multifaceted growth, as well as practical constraints such as limited time, funding, and research capacity. Therefore, to fully understand the impact of free play, it is essential to adopt a comprehensive analysis approach that captures its contributions across multiple areas of child development.

\subsection{Impact of Physical Setting on Play Effectiveness}
Physical settings, which encompass the tangible, structural, and spatial characteristics of an environment, are essential for supporting rich play experiences and providing opportunities that align with children's play preferences \cite{jasem2022social,van_liempd_young_2018}. Well-designed spaces and diverse materials have been shown to address challenges posed by limited teacher guidance during play.

A substantial body of evidence highlights the multifaceted impact of physical settings on children's development. \cite{li_development_2023} showed that outdoor free play enhances physical fitness, learning approaches, social interaction, and imagination. Studies providing playgrounds and sports equipment \cite{tandon_two_2019,ruiz2020analysis,tortella_effects_2022} found consistent benefits for motor skills, including increased activity intensity and improved coordination. \cite{mccree2020hare} emphasized that natural play settings support emotional well-being, while \cite{kukkonen_creative_2020} noted that open-ended environments, diverse materials, and intrinsic motivation foster creativity and collaboration. Together, these findings show that physical settings support children’s cognitive, motor, emotional, and social development.

Despite these insights, our understanding of how different play settings shape children's development across multiple domains remains limited. While existing research has demonstrated the benefits of specific environments—such as natural settings for emotional regulation or playgrounds for motor skill development—there remains a paucity of studies examining how children's developmental trajectories and performance outcomes vary across diverse play settings. 

\subsection{Assessment of Children's Development}
Free play is a spontaneous, self-initiated, and self-regulated activity that is relatively risk-free and not necessarily goal-oriented \citep{verenikina2003child}. This unstructured nature poses challenges for evaluation, as it hinders the establishment of clear assessment criteria. Moreover, variability in play behaviors and the subjective nature of play further complicate standardized outcome measurement.

Empirical studies have adopted diverse methods to assess development in the context of free play. \cite{colliver2022free} used time-use diaries and composite self-regulation scores from parent, teacher, and observer reports to examine the predictive link between early play and later self-regulation. \cite{abe2023one} tracked changes in handgrip strength and forearm muscle thickness over a year in 111 kindergarteners, grouped by teacher-reported play preferences. \cite{zeng2024free} employed a teacher action research approach, combining video/audio recordings, observation checklists, and teacher journals to explore how open-ended questions during Loose Parts Play supported science process skills and independent inquiry.

Overall, traditional methods for evaluating children’s development have primarily relied on human observation and standardized tests, both of which have notable limitations. Human observation is time-consuming and highly dependent on the observer’s expertise, making it susceptible to bias and inconsistency. It also often fails to capture children’s internal thoughts and emotional experiences. Standardized tests, while offering greater objectivity, are typically better suited for assessing measurable motor skills and may inadequately reflect emotional and social abilities.

\subsection{The Present Study}
\label{sec_lr_ability}

To address the gap outlined above, this study proposes an innovative approach to comprehensively assess children's abilities across multiple developmental domains. We introduce an experimental ability framework based on the Early Childhood Development Index 2030 (ECDI2030) \citep{ecdi2030}, which includes learning (e.g., early numeracy, language, communication), psychosocial well-being (social and emotional development), and health (self-care, gross motor skills). We redefine several representative abilities from the cognitive, motor, emotional, and social domains (see Table \ref{table_ability_def}). 

\begin{table*}[htbp!]
    \centering
    \caption{Abilities involved in child development}
    \fontsize{10pt}{12pt}\selectfont 
    \begin{tabular}{ p{1.5cm} p{2cm} p{11.5cm}}
        \hline
         \textbf{Domain} & \textbf{Ability Dimension} & \textbf{Definition} \\
         \hline
         \multirow{2}{1.5cm}{Cognitive} &  Numeracy and geometry & The ability to understand and manipulate numbers, shapes, spatial relationships, and patterns. It encompasses the development of number sense, mathematical logical reasoning, the recognition and creation of geometric shapes, as well as the understanding of objects' positions and orientations in space. \\
          &  Creativity and imagination & The ability to generate novel ideas and unique solutions. This includes creative thinking, exploration of artistic expressions, story creation, and the tendency to use everyday objects in unconventional ways.\\
         \hline
         \multirow{2}{1.5cm}{Motor} &  Fine motor development & The development of the small muscle groups in the hands and fingers, enabling children to perform delicate actions such as holding a pen, buttoning, cutting paper, and drawing. \\
          &  Gross motor development & The development of coordination and strength in the large muscle groups, as demonstrated in activities such as running, jumping, climbing, throwing, and catching. It is crucial for maintaining balance, posture control, and overall physical activity. \\
         \hline
         \multirow{2}{1.5cm}{Emotional} & Emotion recognition & The ability to recognize, understand, and label one's own and others' emotions. It includes the awareness of basic emotions such as happiness, sadness, anger, and fear, as well as the understanding of more complex emotions as one grows older. \\
          &  Empathy & The ability to understand and share the feelings of others, involving the recognition of others' emotional states and making appropriate social responses based on these states. It promotes prosocial behavior and the establishment of social relationships. \\
         \hline
         \multirow{2}{1.5cm}{Social} & Communication & The ability to express thoughts and emotions through language (both verbal and written), non-verbal signals (such as facial expressions and gestures), and symbolic systems. This includes vocabulary knowledge, understanding and use of grammatical structures, and the social rules of effective communication. \\
          &  Collaboration & The ability of individuals to work collaboratively with others towards a common goal. This includes sharing resources, taking turns, resolving conflicts, and following group rules. It fosters skills in teamwork and collective problem-solving. \\ 
         \hline
    \end{tabular}
    \label{table_ability_def}
\end{table*}

Using this framework, we developed an LLM-based approach to analyze children's developmental performance through their data. We collected daily self-narratives from a kindergarten class over one semester, describing children's own play experiences across four different free play settings. To evaluate the effectiveness of the proposed approach, we addressed the following two research questions:

\begin{itemize}
    \item[ ] \textit{RQ1. How reliable is the LLM-based approach in analyzing child development through self-narratives of play experiences?}
    \item[ ] \textit{RQ2. To what extent can the LLM-based approach reflect children's development across different ability dimensions in free play across various physical settings?}
    \begin{itemize}
        \item[ ] \textit{RQ2.1. How do children perform across different ability dimensions?}
        \item[ ] \textit{RQ2.2. How do children's developmental outcomes differ in different play settings?}
    \end{itemize}
\end{itemize}

\section{The LLM-Based Approach for Analyzing Children’s Development}
\label{sec_apch}
This section details the design of the LLM-based approach for analyzing child development and includes illustrative examples from real-world scenarios to showcase its application and functionality.

\subsection{Technical Framework}
\begin{figure*}[htbp!]
    \centering
    \includegraphics[scale=0.38]{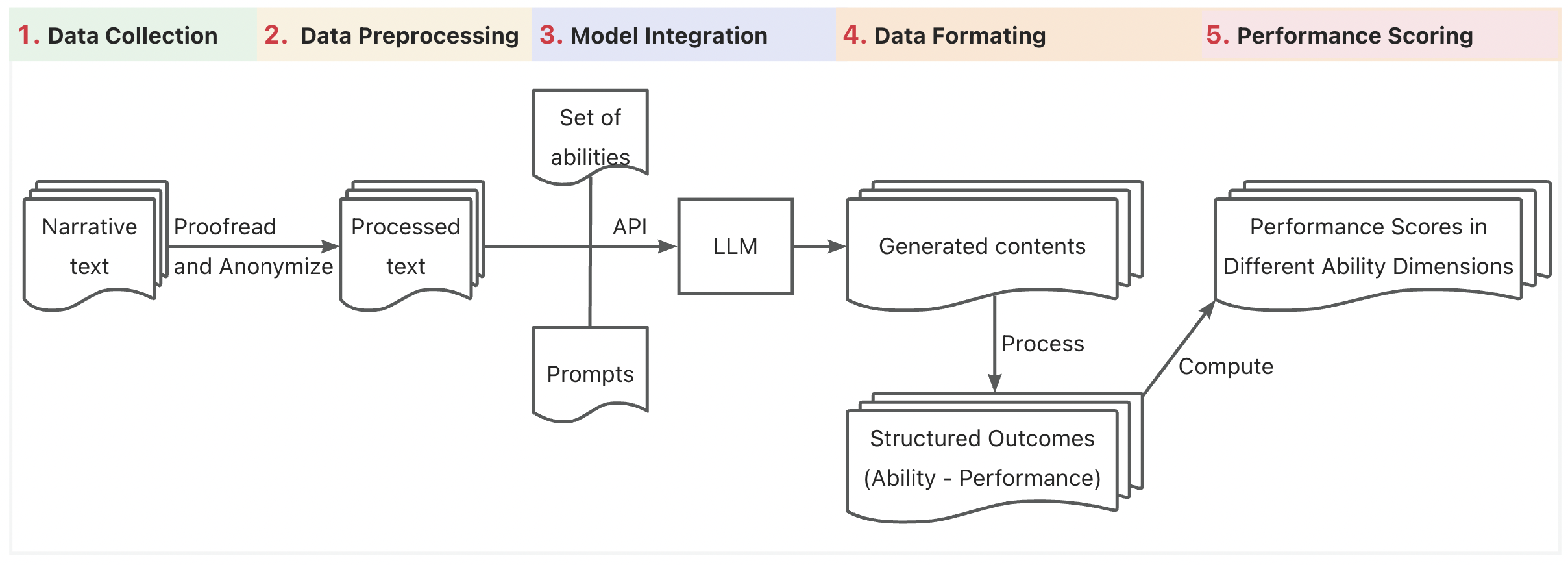}
    \caption{Technical framework of the LLM-base approach} 
    \small
    \label{fig_frwk}
 \end{figure*}

Figure \ref{fig_frwk} presents the technical framework of the proposed approach, which consists of five key steps: Data Collection, Data Preprocessing, Model Integration, Data Formatting, and Ability Scoring.

\begin{itemize}
    \item \textbf{Data Collection}: Children's narrative texts about free play are collected, and a database is designed to store this data. The specific data design is given in the next section.
    \item \textbf{Data Preprocessing}: Children's narrative texts are subjected to proofreading, which includes correcting spelling errors. Furthermore, to protect their privacy, any names or nicknames of children that appear in the narratives are replaced with unique identifiers.
    \item \textbf{Model Integration}: By invoking the LLM API, predefined prompts and a set of abilities related to various aspects of children’s development, along with the processed narrative texts, are input. The next section will provide the specific prompt design.
    \item \textbf{Data Formatting}: The contents generated by the LLM are processed into structured outcomes in a format that associates abilities with performance. Detailed explanations of this data design will be provided in the next section.
    \item \textbf{Performance Scoring}: The performance scores are calculated based on the structured outcomes from the previous step. The calculation formula for a particular ability is the number of times that ability is inferred within a certain period, divided by the total number of activity records for that child during the same period. Detailed explanations of this data design will be provided in the next section.
\end{itemize}

\subsection{Data Design} \label{subsec_data_design}
The data structure is designed to describe children from two aspects: 1) \textit{What, when, and where did children play?} 2) \textit{How did they perform in different ability dimensions?} We utilize a relational database to store the data, with one entity table and two relational tables designed. The definitions and structures of all tables are listed in Table \ref{table_data_design}. 

\begin{table*}[htbp!]
    \centering
    \caption{Data design}
    \fontsize{10pt}{12pt}\selectfont 
    \begin{tabular}{ p{3cm} p{12cm}}
    \hline
    \textbf{Table} & \textbf{Description} \\
    \hline
    Child & An entity table that stores the personal information of children, including children' unique ID, name, nickname, year of birth, gender, and class information they belong to. \\
    Child-Activity & A relational table that stores data related to the activity, including the unique ID of the activity, the unique ID of the child who engaged in the activity, the text narrated by the child, the processed narrative text, the location where the activity happened, and the date when the activity happened. \\
    Ability-Performance & A relational table that stores children' performance analyzed by the proposed approach, including the unique IDs of the activity and the child, the ability inferred from the child's narrative, the description of the performance analyzed from the narrative that supports the inference. \\
    Student-Ability & A relational table that stores children' performance scores in different ability dimensions, including the unique ID of the child, the location where the activity happened, the ability, the performance score, the start and end date of the activities used for calculation (typically measured in weeks, months, or semesters) \\
    \hline
    \end{tabular}
    \label{table_data_design}
\end{table*}

    


 
\subsection{Prompts Design}\label{subsec_prmpt_design}
Prompt engineering is a pivotal technique for enhancing the versatility and performance of LLMs by strategically designing task-specific prompts that guide model behavior without altering the underlying parameters \citep{chen2023unleashing,sahoo2024systematic}. Following the strategies recommended in \cite{openai2023}'s guide, we adopt the following strategies in designing our prompts: (1) Ensure tasks are clear and specific, (2) Use separators for distinction, (3) Employ a framework architecture in writing prompts, (4) Request structured output, (5) Iteratively test and refine prompts.  In this study, we adopted the abilities defined in Table \ref{table_ability_def}. After several rounds of testing and modifications, we ultimately determined the following structured prompts: 

\begin{itemize}
    \item[ ] \textit{\textbf{Role:} You are an expert in the field of early childhood education.}
    \item[ ]  \textit{\textbf{Task Description:} Based on a child's self-narrative of their play process, select the most relevant and prominent abilities from the following [Ability Categories] and summarize the child's behavior in relation to these abilities.}
    \item[ ]  \textit{\textbf{Ability Categories: }``Numerical and Geometric Cognition", ``Creativity and Imagination", ``Fine Motor Development", ``Gross Motor Development", ``Empathy", ``Emotional Cognition", ``Communication, ``Cooperation".}
    \item[ ]  \textit{\textbf{Requirements:}}
    \begin{enumerate}
        \item \textit{Select only the most relevant and prominent abilities, not exceeding the given range of abilities.}
        \item \textit{Provide a brief and specific description of the child's behavior for each ability, with the child as the subject.}
        \item \textit{Format the output in table form, with each row representing an ability and its corresponding behavior.}
        \item \textit{Each row in the table should include two columns: Ability and Behavior.}
        \item \textit{No additional explanations or extra information is needed, only a summary of the abilities and behaviors.}
    \end{enumerate}
    \item[ ]  \textit{\textbf{Input Data:} (Child's self-narrative text)}
\end{itemize}

\section{Research Method}
\label{sec_mtd}
Within the scope of the research process, from the design of the LLM-based analysis approach to the evaluation procedures, we collected extensive data from children, analyzed it using the proposed approach, and evaluated the reliability through a single-group study design, incorporating both qualitative and quantitative feedback on the analysis outcomes.

\subsection{Contexts and Participants}
This study was conducted over a semester in a public kindergarten in Zhejiang Province, China, from September 4, 2023, to January 24, 2024. 
The kindergarten implements the Anji Play teaching model \citep{coffino2019anji}, which prioritizes child-led, play-based learning, fostering opportunities for children to explore, create, and interact meaningfully with their environment. The kindergarten provides four different types of areas for free play (see Table \ref{table_area}).

\begin{table*}[htbp!]
    \centering
    \caption{Physical Settings for Free Play}
    \fontsize{10pt}{12pt}\selectfont 
    \begin{tabular}{ p{3.4cm} p{5.8cm} p{5.8cm}}
        \hline
         \textbf{Area} & \textbf{Features} & \textbf{Materials} \\
         \hline
            Sand-water Area & An outdoor sandy area & Faucets, plastic hoses, buckets, shovels, and wooden molds \\
            Hillside-zipline Area & A grassy field with a small hill, a running track surrounding the hill, a rope slop and a slide & Ladders, planks, plastic pipes, soft cushions, plastic sleds, tires, yoga balls and children's scooters \\
            Building Blocks Area & An outdoor concrete surface & Wooden building blocks \\
            Playground Area & A large outdoor concrete surface & Balls, tires, planks, ladders, plastic pipes, and children's bicycles. \\
         \hline
    \end{tabular}
    \label{table_area}
\end{table*}

The open-ended materials enable children to explore freely without external intervention. Each class is assigned to different play areas on a monthly rotation to ensure diverse experiences, with play activities relocated to indoor venues on rainy days.

Twenty-nine healthy children, aged 5 to 6 years (comprising 13 females and 16 males), along with their two teachers, agreed to participate in the study. 
The evaluation process involved eight professionals, including five master's students with bachelor's degrees in early childhood education and teaching qualifications, one associate professor and two assistant researchers with extensive research experience in the field of education. Before the evaluation, a thirty-minute training session was conducted to explain the evaluation process and ensure consistency in the assessments.

\subsection{Study Design and Procedure}
\label{subsec_pcxt}
The study was approved by the Research Ethics Committee of the authors' affiliated institution. All the participants, their parents and teachers were explained about the project and the collected data for the research. Written consent was obtained from them. 

First, as a daily routine, children engaged in one hour of free play in the morning. Teachers were responsible for leading children to the assigned areas and offering help when required without interfering in their play. After the playtime, children voluntarily drew what they played that day. Then, the teachers conducted one-to-one interviews with children, requiring them to narrate what they played that day based on their drawings. Teachers used a speech-to-text tool to record their narratives and saved the texts after proofreading. 

Next, the proposed approach was applied to analyze each narrative, generating multiple analysis outcomes. Then, we employed a simple random sampling \citep{cochran1977sampling} to assess the analysis outcomes. 
Each professionals evaluated a random selection part of the samples, without duplication. Each item contains a narrative and several outcomes of children's \textbf{Ability-Performance}. For each outcome, the professionals answered yes-or-no questions regarding the following: 

For those identified abilities: 
\begin{itemize}
    \item \textbf{Semantic Consistency}: Whether the generated description of the performance is consistent with the content of the child's narrative?
    \item \textbf{Ability Relevance}: Whether the generated description of the performance matches with the identified ability?
\end{itemize}

For those unidentified abilities:
\begin{itemize}
    \item \textbf{Identification Omission}: Whether the child's narrative reflects an ability, but AI failed to identify it?
\end{itemize}

After completing all the items, the professionals were asked to answer two additional open-ended questions based on their process of filling out the questionnaire.

\begin{itemize}
    \item What do you think about the advantages of this technology?
    \item What do you think about the drawbacks of this technology?
\end{itemize}

\subsection{Data Collection and Analysis}
\label{subsec_col}
Children's personal information (including name, gender, and age), the schedule of free play, and children's narratives were obtained from their teachers. A total of 2,224 narratives from 29 children were collected over 95 school days in five months. The descriptive statistics of the number of children's narratives are as follows: the mean is 76.67, the variance is 13.05, the minimum is 49, and the maximum is 94. The observed variance could be attributed to factors such as occasional absences due to illness or personal reasons, which may have led to some children missing school days and thus affecting the overall data. 

The qwen-max model \citep{bai2023qwen} is used in the proposed approach. With the input of all narrative texts, 11,389 analysis outcomes of children's \textbf{Ability-Performance} were generated and saved in the Ability-Performance Table. Next, a simple random sampling \citep{cochran1977sampling} was employed to evaluate the reliability of the proposed approach. We adopted the standard formula for sample size determination and randomly selected a total of 328 samples from the 2,224 records for this study. Utilizing Python for the execution of the proposed approach, we successfully processed the dataset and obtained 2,624 analysis outcomes. Eight professionals collaboratively completed the evaluation questionnaires for all outcomes. Furthermore, based on their experience of evaluation, they offered opinions regarding the proposed approach, with 21 comments highlighting advantages and 24 addressing drawbacks. We manually coded these comments into several categories. 

The ability scores of 29 children across eight abilities in four areas over one semester were then computed based on the analysis outcomes. Based on the mean values, we constructed radar charts for the ability scores in each area. Subsequently, we conducted the Shapiro-Wilk test to assess whether the data followed a normal distribution. Depending on the test results, we applied the Kruskal-Wallis H test for non-normally distributed data and ANOVA for normally distributed data to examine the differences in each ability across different areas. Finally, based on the significance of the results, we performed post hoc analysis to further clarify the differences among groups.

\section{Results}
\label{sec_rslt}
\subsection{Reliability of the Analysis Outcomes Generated from LLM}
\label{sec_rslt1}
Based on the results of the questionnaire-based survey, we computed statistics on both identified abilities and unidentified abilities. For identified abilities, we evaluated the analysis outcomes in terms of Semantic Consistency and Ability Relevance, as well as Accuracy, which is counted when both Semantic Consistency and Ability Relevance are true. For unidentified abilities, we computed the Identification Omission.

\subsubsection{Semantic Consistency, Ability Relevance and Accuracy}
We counted all identified abilities (\textbf{Total Count}). The computation of \textbf{Semantic Consistency} and \textbf{Ability Relevance} involves the proportion of items that are affirmatively responded with `yes' relative to the \textbf{Total Count}. \textbf{Accuracy} is determined by calculating the proportion of items that are affirmatively responded with `yes' for both \textbf{Semantic Consistency} and \textbf{Ability Relevance} relative to the \textbf{Total Count}. The result is presented in Table \ref{table_rlt_accuracy}. 

\begin{table*}[htbp!]
    \centering
    \caption{Evaluation result of the proposed approach on the test sample}
    \fontsize{10pt}{12pt}\selectfont 
    \begin{tabular}{ lcccc }
        \hline
        \textbf{Ability Count} & \textbf{Total Consistency} & \textbf{Semantic Consistency}  & \textbf{Ability Relevance}  & \textbf{Accuracy} \\
        \hline
        All & 1714 & 93.3\% & 94.1\% & 91.1\% \\
        Numerical and Geometric Cognition & 122 & 95.1\% & 95.9\% & 92.6\% \\
        Creativity and Imagination & 321 & 96.6\% & 96.0\% & 95.0\% \\
        Fine Motor Development & 248 & 92.7\% & 91.9\% & 90.7\% \\
        Gross Motor Development & 139 & 94.2\% & 100.0\% & 90.6\% \\
        Emotional Recognition & 215 & 88.4\% & 86.5\% & 84.2\% \\
        Empathy & 52 & 76.9\% & 75.0\% & 73.1\% \\
        Communication & 324 & 92.3\% & 96.0\% & 90.1\% \\
        Cooperation & 293 & 96.6\% & 97.3\% & 96.2\% \\
        \hline
        \multicolumn{5}{l}{\textit{The sample size for the narratives is 328, with a total of 2,624 generated analysis outcomes.}} \\
    \end{tabular}
    \label{table_rlt_accuracy}
\end{table*}

Overall, Semantic Consistency, Ability Relevance, and Accuracy all exceeded 90\%, representing a high reliability. Regarding the results for each ability, Semantic Consistency, Ability Relevance, and Accuracy for Creativity and Imagination, Fine Motor Development, Gross Motor Development, Communication, and Cooperation all surpassed 90\%. However, Emotional Recognition and Empathy scored relatively lower, with values just above 80\% and 70\%, respectively. Furthermore, the number of outcomes for Gross Motor Development and Empathy did not exceed half of the total test sample size, while all other abilities surpassed this threshold. 

\subsubsection{Identification Omission}
We counted all unidentified abilities (\textbf{Total Count}) and items that are affirmatively responded with `yes' (\textbf{Count of Omissions}). The computation of \textbf{Omission rate} involves the proportion of \textbf{Count of Omissions}  relative to the \textbf{Total Count}. The result is presented in Table \ref{table_rlt_omission}. 

\begin{table*}[htbp!]
    \centering
    \fontsize{10pt}{12pt}\selectfont 
    \caption{Identification omission of the proposed approach on the test sample}
    \begin{tabular}{ lccc }
        \hline
        \textbf{Ability}  & \textbf{Total Count} & \textbf{Count of Omissions} & \textbf{Omission Rate} \\
        \hline
        All & 910 & 128 & 14.1\% \\
        Numerical and Geometric Cognition & 206 & 44 & 21.4\% \\
        Creativity and Imagination & 7 & 1 & 14.3\% \\
        Fine Motor Development & 80 & 12 & 15.0\% \\
        Gross Motor Development & 189 & 50 & 26.5\% \\
        Emotional Recognition & 113 & 2 & 1.8\% \\
        Empathy & 276 & 15 & 5.4\% \\
        Communication & 4 & 1 & 25.0\% \\
        Cooperation & 35 & 3 & 8.6\% \\
        \hline
        \multicolumn{4}{l}{\textit{The sample size for the narratives is 328, with a total of 2,624 generated analysis outcomes.}}\\
    \end{tabular}
    \label{table_rlt_omission}
\end{table*}

Overall, the method's rate of Identification Omission reached 14.1\%. Specifically, the omission rates for Numerical and Geometric Cognition, Gross Motor Development, and Communication exceeded 20\%, while for Creativity and Imagination, Fine Motor Development, the omission rates were above 10\%. In terms of Emotional Recognition, Empathy, and Cooperation, the omission rates were below 10\%. This indicates that the proposed approach still omits the inference of abilities to some extent. 

\subsubsection{Comments of Professionals on the Open-ended Questions}
In the open-ended questions regarding the advantages of the technology, we collected 21 comments, categorized into four dimensions: \textbf{accuracy}, \textbf{efficiency}, \textbf{ease of use}, and \textbf{value}. Firstly, ten comments emphasized the accuracy of outcomes, noting the system’s ability to reliably extract and classify abilities such as numerical cognition, fine motor development, creativity, and imagination. It was praised for its consistency and accurate interpretation of children’s narratives. Secondly, six comments highlighted efficiency, pointing out that the technology significantly reduces the workload of analyzing large volumes of text, offering a more convenient and scalable alternative to manual assessment. Thirdly, two comments mentioned its ease of use, describing it as user-friendly and easy for ordinary teachers to adopt without needing advanced technical skills. Lastly, four comments focused on its value, recognizing its potential to reveal overlooked abilities—especially in the social domain—and its practical applications in building digital profiles and identifying developmental patterns to support teaching and parenting.

In the open-ended questions regarding the drawbacks of the technology, we collected 24 comments—20 pointed out limitations, and 4 offered suggestions for improvement. These limitations can be categorized into five dimensions: (1) \textbf{Misinterpretation of Activities and Abilities}: The AI may inaccurately infer abilities based on participation in certain activities, assuming the child has the ability without sufficient evidence, leading to potential misjudgments. (2) \textbf{Ambiguity in Ability Definitions}: Some ability definitions are too abstract or overlapping (e.g., emotional cognition vs. empathy), making accurate classification challenging for the AI. (3) \textbf{Language and Expression Challenges}: Children's language can be vague or ambiguous, and the AI struggles with understanding emotional context, tone, and expressions, affecting analysis accuracy. (4) \textbf{Overinterpretation and Subjectivity}: Especially in emotional domains, the AI may infer abilities not clearly stated, leading to overinterpretation and reduced objectivity. (5) \textbf{Technical Limitations}: Spelling errors, complex emotions, and standardized content can hinder the AI’s performance, potentially overlooking individual differences and creativity.

Moreover, professionals suggested that relying solely on children's game narratives as the data source is insufficiently comprehensive. Inferring children's behaviors and abilities based only on their self-narratives can be somewhat one-sided. Given the limited expressive capabilities of children, it is necessary for teachers or researchers to further verify the text analyzed by AI, which still requires additional time. Additionally, aligning the analysis with guidelines and frameworks can allow for more refined judgments of children's developmental abilities across different age groups.

\subsection{Children's Performance in Different Ability Dimensions}
\label{sec_rslt2}
We present a comprehensive analysis of the ability scores across different play areas for 29 children. Table \ref{table_descriptive} provides descriptive statistics of these ability scores, including the mean, standard deviation (SD), as well as the results from the Shapiro-Wilk test, which are represented by the W statistic and the corresponding P-value. Additionally, the table shows the range (minimum and maximum) for each ability within the four areas: Building Blocks Area, Sand-water Area, Hillside-zipline Area, and Playground Area. 

\begin{table*}[htbp!]
    \centering
    \caption{Descriptive statistics of ability scores (N=29)}
    \fontsize{10pt}{12pt}\selectfont 
    \begin{tabular}{ p{4cm} p{1.5cm} p{1.5cm} p{1.5cm} p{1.5cm} p{2cm} }
        \hline
        \textbf{Ability} & \textbf{Mean} & \textbf{SD} & \textbf{W} & \textbf{P-Value} & \textbf{Min-Max} \\
        \hline
        \textit{Building Blocks Area} & & & & & \\
        \hline
        Numeracy and Geometry & 0.493  & 0.150  & 0.947  & 0.152  & 0.143-0.737 \\
        Creativity and Imagination & \textbf{0.990}  & 0.018  & 0.622  & $<.001$ & 0.944-1.000 \\
        Fine Motor Development & \textbf{0.791}  & 0.082  & 0.966  & 0.458  & 0.649-0.950 \\
        Gross Motor Development & \textcolor{red}{0.298}  & 0.099  & 0.954  & 0.230  & 0.143-0.529 \\
        Emotion Recognition & \textcolor{red}{0.311}  & 0.137  & 0.920  & 0.030  & 0.122-0.714 \\
        Empathy & \textcolor{red}{0.162}  & 0.102  & 0.969  & 0.520  & 0.000-0.382 \\
        Communication & \textcolor{red}{0.209}  & 0.193  & 0.770  & $<.001$ & 0.000-1.000 \\
        Collaboration & \textbf{0.903}  & 0.071  & 0.895  & 0.007  & 0.684-1.000 \\
        \hline
        \textit{Sand-water Area} & & & & & \\
        \hline
        Numeracy and Geometry & 0.379  & 0.160  & 0.943  & 0.124  & 0.154-0.700 \\
        Creativity and Imagination & \textbf{0.969}  & 0.069  & 0.518  & $<.001$ & 0.667-1.000 \\
        Fine Motor Development & \textbf{0.823}  & 0.136  & 0.938  & 0.086  & 0.500-1.000 \\
        Gross Motor Development & \textcolor{red}{0.265}  & 0.159  & 0.971  & 0.591  & 0.000-0.600 \\
        Emotion Recognition & 0.680  & 0.194  & 0.940  & 0.101  & 0.167-1.000 \\
        Empathy & \textcolor{red}{0.234}  & 0.203  & 0.906  & 0.013  & 0.000-0.750 \\
        Communication & \textbf{0.996}  & 0.017  & 0.288  & $<.001$ & 0.923-1.000 \\
        Collaboration & \textbf{0.926}  & 0.126  & 0.656  & $<.001$ & 0.500-1.000 \\
        \hline
        \textit{Hillside-zipline Area} & & & & & \\
        \hline
        Numeracy and Geometry & \textcolor{red}{0.173}  & 0.162  & 0.846  & $<.001$ & 0.000-0.688 \\
        Creativity and Imagination & \textbf{0.886}  & 0.073  & 0.940  & 0.100  & 0.688-1.000 \\
        Fine Motor Development & 0.458  & 0.144  & 0.980  & 0.849  & 0.200-0.786 \\
        Gross Motor Development & \textbf{0.721}  & 0.136  & 0.866  & 0.002  & 0.357-0.923 \\
        Emotion Recognition & \textbf{0.760}  & 0.185  & 0.915  & 0.023  & 0.313-1.000 \\
        Empathy & \textcolor{red}{0.130}  & 0.095  & 0.926  & 0.043  & 0.000-0.357 \\
        Communication & \textbf{0.970}  & 0.044  & 0.699  & $<.001$ & 0.846-1.000 \\
        Collaboration & \textbf{0.835}  & 0.157  & 0.890  & 0.006  & 0.500-1.000 \\
        \hline
        \textit{Playground Area} & & & & & \\
        \hline
        Numeracy and Geometry & \textcolor{red}{0.283}  & 0.145  & 0.979  & 0.824  & 0.000-0.615 \\
        Creativity and Imagination & \textbf{0.983}  & 0.035  & 0.540  & $<.001$ & 0.875-1.000 \\
        Fine Motor Development & \textbf{0.715}  & 0.151  & 0.970  & 0.570  & 0.455-1.000 \\
        Gross Motor Development & 0.633  & 0.155  & 0.954  & 0.234  & 0.375-1.000 \\
        Emotion Recognition & 0.604  & 0.249  & 0.933  & 0.067  & 0.000-0.947 \\
        Empathy & \textcolor{red}{0.160}  & 0.141  & 0.909  & 0.016  & 0.000-0.500 \\
        Communication & \textbf{0.990}  & 0.028  & 0.426  & $<.001$ & 0.882-1.000 \\
        Collaboration & \textbf{0.913}  & 0.139  & 0.685  & $<.001$ & 0.400-1.000 \\
        \hline
    \end{tabular}
    \label{table_descriptive}
\end{table*}

Based on the Shapiro-Wilk test results, the ability scores in the Building Blocks Area for Numeracy and Geometry, Fine Motor Development, Gross Motor Development, and Empathy; in the Sand-water Area for Numeracy and Geometry, Fine Motor Development, Gross Motor Development, and Emotion Recognition; in the Hillside-zipline Area for Creativity and Imagination, and Fine Motor Development; in the Playground area for Numeracy and Geometry, Fine Motor Development, Gross Motor Development, and Emotion Recognition were found to be normally distributed, as indicated by P-values above 0.05. All other ability scores across the different areas exhibited non-normal distributions, with P-values below the significance threshold of 0.05, suggesting that the data sets were non-normal.

\begin{figure*}[htbp!]
    \centering
    \includegraphics[scale=0.3]{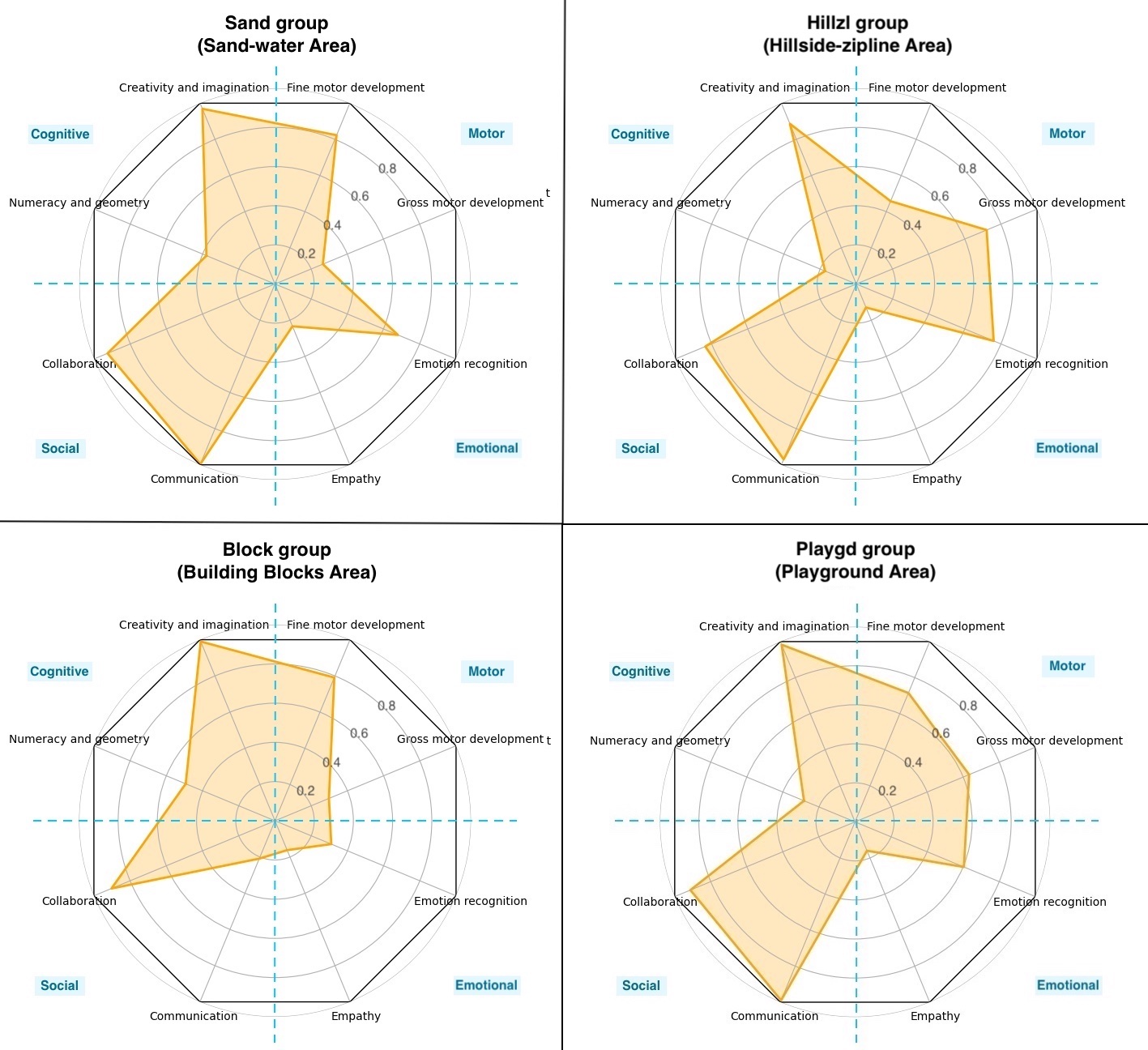}
    \caption{Children's performance across different ability dimensions} 
    \label{fig_perform}
\end{figure*}

The radar charts (Figure \ref{fig_perform}) visually represent the mean ability scores for each area, highlighting the distribution and relative strengths across eight abilities: Numeracy and Geometry, Creativity and Imagination, Fine Motor Development, Gross Motor Development, Emotion Recognition, Empathy, Communication, and Collaboration. Based on our adopted equal-interval segmentation method, scores are categorized as follows: low for scores ranging from 0.0 to 0.33, moderate for scores between 0.34 and 0.66, and high for scores from 0.67 to 1. This segmentation allows us to clearly delineate the performance levels of children across different play areas. 

The Sand-water Area showed high ability levels for Creativity and Imagination, Fine Motor Development, Emotion Recognition, Communication, and Collaboration, with Numeracy and Geometry at a moderate level. Gross Motor Development and Empathy were rated low.

In the Hillside-zipline Area, Creativity and Imagination, Gross Motor Development, Emotion Recognition, Communication, and Collaboration were rated high, with Fine Motor Development at a moderate level. Numeracy and Geometry, and Empathy were rated low.

In the Building Blocks Area, children demonstrated high levels in Creativity and Imagination, Fine Motor Development, and Collaboration. Numeracy and Geometry was at a moderate level, while Gross Motor Development, Emotion Recognition, Empathy, and Communication were at low levels.

For the Playground Area, high ability levels are observed in Creativity and Imagination, Fine Motor Development, Communication, and Collaboration. Gross Motor Development and Emotion Recognition falls into the moderate category, while Numeracy and Geometry, and Empathy are rated low.

These findings suggest that each play area may uniquely contribute to the development of specific abilities in children, highlighting the varying developmental benefits associated with different play settings.

\subsection{Difference of Children's Performance in Various Physical Settings}
\label{sec_rslt3}
The Shapiro-Wilk test results indicated that only the scores for Fine Motor Development were normally distributed across all four settings. Consequently, ANOVA was applied to analyze the Fine Motor Development scores, while the Kruskal-Wallis H test was used for the remaining abilities. The analysis results, presented in Table \ref{table_anova}, demonstrate that significant differences exist across the four settings for all abilities except Empathy. 

\begin{table*}[htbp!]
    \centering
    \fontsize{10pt}{12pt}\selectfont 
    \caption{ANOVA/Kruskal-Wallis H test results for ability scores across different settings}
    \begin{tabular}{ l c c c c }
        \hline
        \textbf{Ability} & \textbf{Method} & \textbf{Statistic} & \textbf{P-value} & \textbf{Significant} \\
        \hline
        Numeracy and geometry & Kruskal-Wallis H test & 44.845 & $<0.001$ & Yes \\
        Creativity and imagination & Kruskal-Wallis H test & 47.109 & $<0.001$ & Yes \\
        Fine motor development & ANOVA & 46.343 & $<0.001$ & Yes \\
        Gross motor development & Kruskal-Wallis H test & 79.639 & $<0.001$ & Yes \\
        Emotion recognition & Kruskal-Wallis H test & 50.349 & $<0.001$  & Yes\\
        Empathy & Kruskal-Wallis H test & 4.561 & 0.207 & No \\
        Communication & Kruskal-Wallis H test & 80.942 & $<0.001$ & Yes \\
        Collaboration & Kruskal-Wallis H test & 11.168 & $0.011$ & Yes \\
        \hline
    \end{tabular}
    \label{table_anova}
\end{table*}

\begin{figure*}[htbp!]
    \centering
    \includegraphics[scale=0.4]{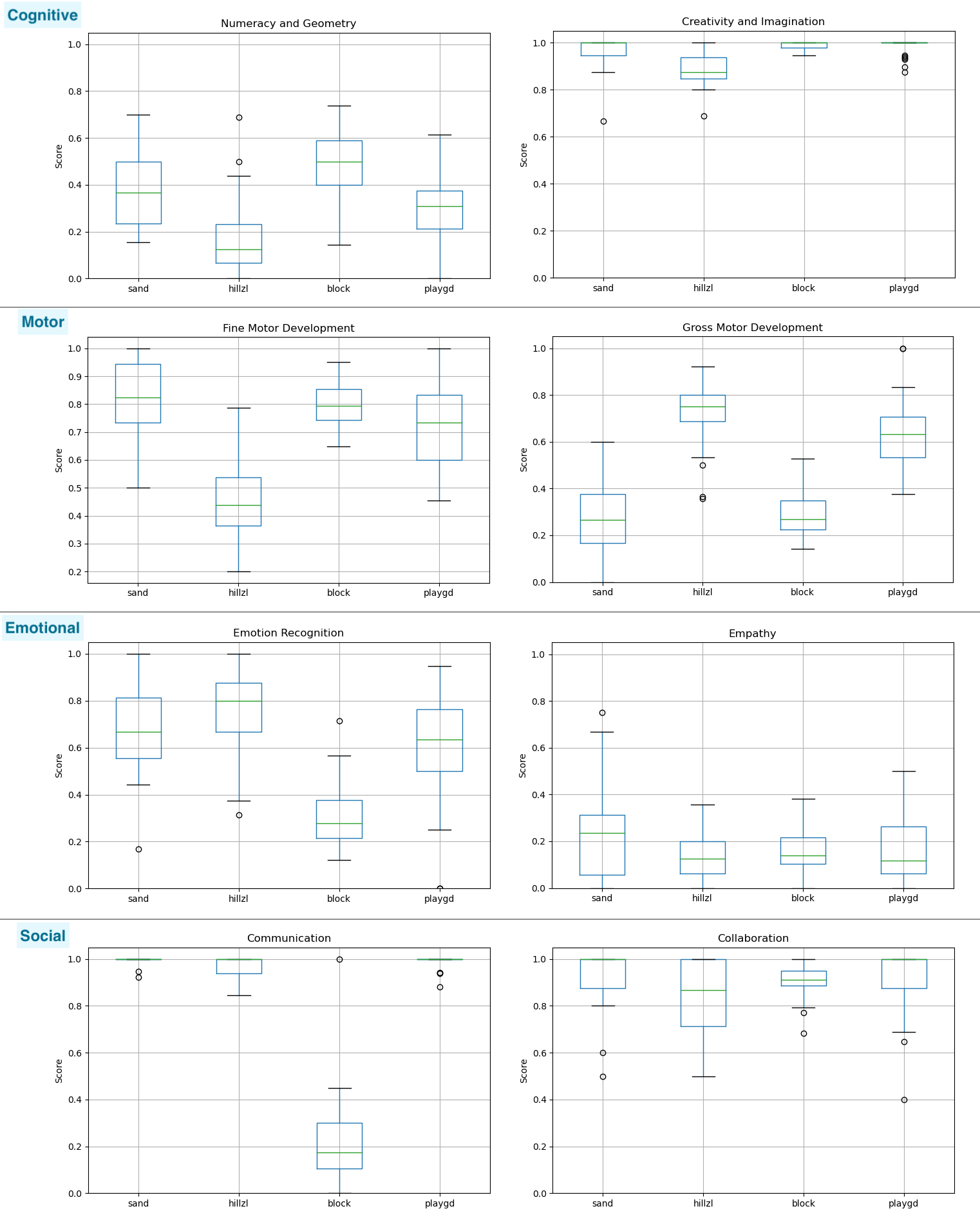}
    \caption{Differences in children's performance across various ability dimensions in different settings}
    \label{fig_diff}
\end{figure*}

Next, box plots were generated to visually illustrate the differences (see Figure \ref{fig_diff}). Then, Tukey's HSD post-hoc analysis was conducted for Fine Motor Development, while Dunn's test was applied for the remaining abilities that exhibited significant differences to further examine intergroup differences. The results are presented in Table \ref{table_post_hoc}. The table shows the pairwise comparisons between the four groups (Sand group for Sand-water Area, Hillzl group for Hillside-zipline Area, Block group for Building Blocks Area, and Playgd for Playground Area), including the mean differences, adjusted p-values, and 95\% confidence intervals. The significant differences are highlighted with "Yes" in the "Significant" column. 

\begin{table*}[htbp!]
    \centering
    \fontsize{10pt}{12pt}\selectfont 
    \caption{Post hoc comparisons of abilities in different settings}
    \begin{tabular}{ lcccc }
        \hline
        \textbf{Comparison} & \textbf{Mean Difference} & \textbf{Adjusted p-value} & \textbf{Confidence Interval (95\%)} & \textbf{Significant} \\
        \hline
        \multicolumn{5}{l}{\textit{Numeracy and geometry}} \\
        \hline  
        Sand vs Hillzl &  -0.206 &  $<.001$  &  [-0.311  , -0.100 ]  &  Yes \\
        Sand vs Block  &  0.114  &  0.029  &  [0.009 , 0.220]  &  Yes \\
        Sand vs Playgd &  -0.096 &  0.089  &  [-0.202  , 0.010]  &  No \\
        Hillzl vs Block  &  0.320  &  $<.001$  &  [0.214 , 0.425]  &  Yes \\
        Hillzl vs Playgd &  0.110  &  0.038  &  [0.004 , 0.215]  &  Yes \\
        Block vs Playgd  &  -0.210 &  $<.001$  &  [-0.316  , -0.104 ]  &  Yes \\
        \hline  
        \multicolumn{5}{l}{\textit{Creativity and Imagination}} \\  
        \hline  
        Sand vs Hillzl &  -0.082 &  $<.001$  &  [-0.119  , -0.046 ]  &  Yes \\
        Sand vs Block  &  0.021  &  0.442  &  [-0.016  , 0.058]  &  No \\
        Sand vs Playgd &  0.015  &  0.734  &  [-0.022  , 0.052]  &  No \\
        Hillzl vs Block  &  0.104  &  $<.001$  &  [0.067 , 0.141]  &  Yes \\
        Hillzl vs Playgd &  0.097  &  $<.001$  &  [0.060 , 0.134]  &  Yes \\
        Block vs Playgd  &  -0.007 &  0.966  &  [-0.044  , 0.030]  &  No \\
        \hline  
        \multicolumn{5}{l}{\textit{Fine Motor Development}} \\
        \hline  
        Sand vs Hillzl &  -0.365 &  $<.001$  &  [-0.455  , -0.275 ]  &  Yes \\
        Sand vs Block  &  -0.032 &  0.794  &  [-0.121  , 0.058]  &  No \\
        Sand vs Playgd &  -0.108 &  0.011  &  [-0.198  , -0.018 ]  &  Yes \\
        Hillzl vs Block  &  0.333  &  $<.001$  &  [0.244 , 0.423]  &  Yes \\
        Hillzl vs Playgd &  0.257  &  $<.001$  &  [0.167 , 0.347]  &  Yes \\
        Block vs Playgd  &  -0.076 &  0.124  &  [-0.166  , 0.013]  &  No \\
        \hline  
        \multicolumn{5}{l}{\textit{Gross Motor Development}} \\   
        \hline  
        Sand vs Hillzl &  0.457  &  $<.001$  &  [0.361 , 0.552]  &  Yes \\
        Sand vs Block  &  0.034  &  0.795  &  [-0.062  , 0.129]  &  No \\
        Sand vs Playgd &  0.368  &  $<.001$  &  [0.273 , 0.464]  &  Yes \\
        Hillzl vs Block  &  -0.423 &  $<.001$  &  [-0.519  , -0.328 ]  &  Yes \\
        Hillzl vs Playgd &  -0.089 &  0.079  &  [-0.184  , 0.007]  &  No \\
        Block vs Playgd  &  0.334  &  $<.001$  &  [0.239 , 0.430]  &  Yes \\
        \hline  
        \multicolumn{5}{l}{\textit{Emotion Recognition}} \\ 
        \hline  
        Sand vs Hillzl &  0.080  &  0.408  &  [-0.054  , 0.214]  &  No \\
        Sand vs Block  &  -0.369 &  $<.001$  &  [-0.503  , -0.235 ]  &  Yes \\
        Sand vs Playgd &  -0.076 &  0.447  &  [-0.210  , 0.057]  &  No \\
        Hillzl vs Block  &  -0.449 &  $<.001$  &  [-0.583  , -0.315 ]  &  Yes \\
        Hillzl vs Playgd &  -0.156 &  0.015  &  [-0.290  , -0.022 ]  &  Yes \\
        Block vs Playgd  &  0.293  &  $<.001$  &  [0.159 , 0.426]  &  Yes \\
        \hline  
        \multicolumn{5}{l}{\textit{Communication}} \\ 
        \hline  
        Sand vs Hillzl &  -0.026 &  0.763  &  [-0.094  , 0.043]  &  No \\
        Sand vs Block  &  -0.786 &  $<.001$  &  [-0.855  , -0.718 ]  &  Yes \\
        Sand vs Playgd &  -0.006 &  0.996  &  [-0.074  , 0.063]  &  No \\
        Hillzl vs Block  &  -0.761 &  $<.001$  &  [-0.829  , -0.692 ]  &  Yes \\
        Hillzl vs Playgd &  0.020  &  0.874  &  [-0.049  , 0.088]  &  No \\
        Block vs Playgd  &  0.780  &  $<.001$  &  [0.712 , 0.849]  &  Yes \\
        \hline  
        \multicolumn{5}{l}{\textit{Collaboration}} \\ 
        \hline  
        Sand vs Hillzl &  -0.091 &  0.037  &  [-0.179  , -0.004 ]  &  Yes \\
        Sand vs Block  &  -0.023 &  0.901  &  [-0.110  , 0.064]  &  No \\
        Sand vs Playgd &  -0.013 &  0.981  &  [-0.100  , 0.075]  &  No \\
        Hillzl vs Block  &  0.068  &  0.180  &  [-0.019  , 0.156]  &  No \\
        Hillzl vs Playgd &  0.079  &  0.094  &  [-0.009  , 0.166]  &  No \\
        Block vs Playgd  &  0.010  &  0.990  &  [-0.077  , 0.098]  &  No \\
        \hline  
    \end{tabular}
    \label{table_post_hoc}
\end{table*}

According to Figure \ref{fig_diff} and Table \ref{table_post_hoc}, it can be observed that for Numeracy and Geometry, significant differences were found among all groups except between the Sand group and the Playgd group. The Block group achieved the highest score, while the Hillzl group had the lowest score. 
For Creativity and Imagination, the Hillzl group scored lower than the other three groups, with no significant differences detected among the latter. 
Regarding Fine Motor Development, the Hillzl group again scored lower than the other three groups, and the Sand group had significantly higher scores than the Playgd group, with no significant differences among the remaining groups. 
For Gross Motor Development, the Hillzl group and the Playgd group had significantly higher scores than the Sand group and the Block group. 
In terms of Emotion Recognition, the Block group scored lower than the other three groups, while the Hillzl group had significantly higher scores than the Playgd group, with no significant differences among the other groups. 
For Empathy, no significant differences were observed among the four groups. 
For Communication, the Block group had significantly lower scores than the other three groups, with no significant differences among the latter. 
For Collaboration, the Sand group had significantly higher scores than the Hillzl group, with no significant differences among the other groups.

\section{Discussion}
\label{sec_dsc}
\subsection{Effectiveness of the LLM-based Approach}

This study proposed an innovative approach that introduced cutting-edge LLM technology. We verified the effectiveness of the proposed approach by addressing two research questions. 

\subsubsection{RQ1: How reliable is the LLM-based approach in analyzing child development through self-narratives of play experiences?}

The proposed approach requires LLM to generate inferences regarding children's ability dimensions, with the evidence for these inferences derived directly from the children's self-narratives about their play experiences. This traceability of evidence to the children's own words helps to ensure a certain level of reliability in the results. Additionally, such an approach provides insights directly from the child's perspective, capturing authentic and firsthand accounts of their experiences. This aligns with the emphasis placed by \cite{fleer2009early} on the importance of respecting children's individual backgrounds and perspectives in early childhood education. 

The results revealed the findings: \textbf{The proposed approach achieved high accuracy ($>$90\%) in identifying children's development in the cognitive, motor, and social domains, while its accuracy in the emotional domain is relatively lower (between 70\% and 90\%)}. The professionals' comments further revealed the impressive consistency and accuracy in categorizing some abilities, as well as understanding the meaning of children's narrative texts. The lower accuracy in inferring the emotional domain can be attributed to the nature of this ability itself, as well as the limitations of the LLM. Cognitive and motor abilities are often more explicitly expressed through actions and outcomes in play narratives, making them easier to identify and analyze. For example, the construction of a tower using blocks clearly indicates fine motor skills and spatial awareness \citep{hill2024influence}. In contrast, emotional abilities such as empathy and emotional recognition are more subtle and context-dependent, requiring a deeper understanding of the narrative's emotional nuances \citep{sailunaz2018emotion}. The LLM may struggle to accurately infer these abilities due to the complexity of emotional expressions and the variability in how children articulate their emotions. Additionally, the training data for LLMs may not be as comprehensive in capturing the subtleties of emotional development compared to cognitive and motor skills. 

Furthermore, the approach still exhibited some omissions in identifying certain abilities. This can be attributed to the inherent limitations of the LLM and the nature of the data. The LLM's performance is heavily reliant on the quality and relevance of the training data \citep{sachdeva2024train}. If the training data lacks sufficient examples of specific abilities or contexts, the model may fail to recognize them in new narratives \citep{schroeder2024can}. Furthermore, the complexity and variability of children's narratives, including differences in language proficiency and narrative style, can also contribute to omissions. Some children may not articulate their experiences in a way that clearly aligns with predefined ability categories, making it difficult for the LLM to accurately infer their abilities.

\subsubsection{RQ2. To what extent can the LLM-based approach reflect children’s development across different ability dimensions in free play across various physical settings?}

Based on the analysis outcomes of children’s daily data over a semester using the LLM, we computed the children’s performance scores across different ability dimensions in various play settings, which is an innovative approach that integrates learning analytics. According to the results in the previous section, we highlight the findings: \textbf{1) Different play settings had distinct impacts on children’s development, with each area uniquely contributing to specific abilities; 2) Significant differences were found across play settings for most ability dimensions, except for empathy.} The radar charts (Figure \ref{fig_perform}) intuitively reflect the performance of eight abilities, while the box plots (Figure \ref{fig_diff}) clearly illustrate the differences in ability development across various play settings. These visualizations are of great help in enabling teachers to monitor the ability differences of children in different play areas. Traditional methods often rely on teachers’ observations, experience, and memory to judge children’s behavior in different play areas over a period of time. In contrast, our approach uses visualizations of children’s data, which can greatly assist teachers in monitoring children’s development, helping them to understand their needs in free play.

Specifically, children exhibited high performance scores in the dimensions of Creativity and Imagination and Collaboration across all play settings, indicating that these abilities can be effectively developed in any of the areas provided. The results also showed that the Building Blocks Area was particularly conducive to the development of Numeracy and Geometry, outperforming the other play areas in this regard. For Fine Motor Development, the Sand-water, Building Blocks, and Playground Area were all supportive of this ability, while the Hillside-zipline Area had a relatively weaker impact. In contrast, Gross Motor Development was more effectively supported by the Hillside-zipline and Playground Area, whereas the Sand-water and Building Blocks Area had less influence on this dimension. Regarding Emotion Recognition and Communication, the Sand-water, Hillside-zipline, and Playground Area were found to be beneficial, while the Building Blocks Area had a relatively weaker effect. However, all play settings showed low performance scores in the dimension of Empathy. These findings demonstrate that the approach can effectively reflect children's development across various ability dimensions in different play settings, clearly revealing the distinct contributions of each setting to specific abilities. 

Therefore, the proposed LLM-based approach is effective in analyzing children's development through their self-narratives of play. It achieves high accuracy in identifying key developmental domains and provides meaningful insights into how different play settings support various abilities. By combining child-centered data with visual analytics, the approach offers a reliable and practical tool for educators to understand and support children's growth in free play contexts.

\subsection{Values and Challenges of the LLM-based Approach}
This automated ability analysis technology is efficient, capable of processing large amounts of data in a short time, and it is user-friendly. Some studies on the application of LLMs have pointed out that under high cognitive load in a complex task, humans are more willing to share workloads with technological aids \citep{skulmowski2023cognitive}. It can be anticipated that LLM-based technologies will be increasingly widely applied in education. Moreover, this technology could help to improve educators' understanding of abstract concepts. \cite{garcia2025reducing} noted by structuring and systematically exploring abstract concepts, generative languages can assist educators in crafting balanced and effective learning experiences. Finally, the value of this technology lies in its potential to bridge the gap between theoretical educational frameworks and practical classroom applications by providing concrete, data-driven insights. As \cite{chiang2024can} pointed out, the advent of generative AI tools, which enable students to easily produce tangible results, calls for a reevaluation of traditional educational practices that have long emphasized achieving predefined results. By combining natural language processing strengths with learning analytics \citep{siemens2011call}, educators could gain deeper insights into children’s developmental trajectories,  and support the development of personalized learning paths tailored to each child’s unique needs \citep{alfirevic2024educational}. These insights enable educators to optimize play environments to better support holistic child development and guide educators in tailoring their instructional methods to meet the diverse developmental needs of children in real-time.

The LLM-based approach, while showing promise in analyzing children's development through self-narratives, faces several significant challenges that need to be addressed. Firstly, the technology's ability to analyze emotional aspects of children's narratives requires enhancement, as the current accuracy in this domain is relatively low. Secondly, there is a need for more precise and clear definitions of the abilities being analyzed to reduce ambiguity and overlap. Thirdly, the technology must improve its understanding of children's language, which is often characterized by its simplicity and lack of clarity, to more accurately interpret the narratives. Finally, the approach cannot avoid the inherent limitations of LLMs, such as LLMs struggle with long inputs,  limiting their use in tasks requiring broad context; LLM outputs are sensitive to prompt variations, making consistent responses difficult \citep{kaddour2023challenges}. These challenges underscore the necessity for further development and refinement of the technology to enhance its reliability and applicability in educational settings.

\subsection{Limitations}
It is important to acknowledge several limitations that may affect the generalizability and applicability of the findings. Firstly, the study’s sample size and geographical location may limit the generalizability of the findings to other educational settings and cultural contexts. Secondly, the study focused solely on children's self-narratives, which, while providing a child-centered perspective, may not capture the full range of developmental aspects. 

\section{Conclusion}
\label{sec_confw}
This study explores the application of Large Language Models (LLMs) to identify child development through self-narratives of free play across different physical settings. The results indicate that the LLM-based approach can reliably identify children's performance in cognitive, motor, and social abilities, with high accuracy in domains such as numerical and geometric cognition, fine motor development, creativity, and collaboration. However, the approach faces challenges in analyzing emotional abilities due to their subtlety and context-dependence. Additionally, by utilizing the approach, this study revealed significant differences in children's development across various play environments and visually demonstrated that each setting makes unique contributions to specific abilities. These findings confirm that the proposed approach for identifying children's development across various free play settings in kindergarten is highly effective. This research highlights the potential of integrating LLMs with learning analytics to provide child-centered insights into developmental trajectories, thereby enhancing educational practices in early childhood education.

\section{Declaration of generative AI and AI-assisted technologies in the writing process}
During the preparation of this work, the authors used ChatGPT in order to check grammar errors and typos. After using this tool/service, the authors reviewed and edited the content as needed and takes full responsibility for the content of the publication.

\bibliographystyle{elsarticle-num}

\end{document}